\documentclass{article}
\usepackage{amsmath,graphicx,mlspconf}
\usepackage[colorlinks=true, linkcolor=blue, citecolor=blue, urlcolor=blue]{hyperref}
\usepackage{xcolor}
\usepackage{pgfplots} % Required for plotting
\usepackage{pgfplotstable}
\usepackage[numbers]{natbib}
\usepackage{tikz}
\usepackage{graphicx}
\usepackage{booktabs}
\usepackage{pgf-pie}
\usepackage{dirtytalk}
\usepackage{enumitem}
\copyrightnotice{979-8-3503-2411-2/25/\$31.00 {\copyright}2025 IEEE}

\toappear{2025 IEEE International Workshop on Machine Learning for Signal Processing, Aug.\ 31-- Sep.\ 3, 2025, Istanbul, Turkey}

\title{Multi-Agent Interactive Question Generation Framework for Long Document Understanding}
\name{%
  \begin{tabular}{c}
    Kesen Wang\thanks{Corresponding author: kwang@humain.ai}, \quad Daulet Toibazar, \quad Abdulrahman Alfulayt, \quad Abdulaziz S. Albadawi, \quad
    Ranya A. Alkahtani, \\ Asma A. Ibrahim, \quad Haneen A. Alhomoud, \quad Sherif Mohamed, \quad Pedro J. Moreno
  \end{tabular}%
}
\address{Humain, Riyadh, KSA}
\begin{document}

\maketitle
\begin{abstract}
Document Understanding (DU) in long-contextual scenarios with complex layouts remains a significant challenge in vision-language research. Although Large Vision-Language Models (LVLMs) excel at short-context DU tasks, their performance declines in long-context settings. A key limitation is the scarcity of fine-grained training data, particularly for low-resource languages such as Arabic. Existing state-of-the-art techniques rely heavily on human annotation, which is costly and inefficient. We propose a fully automated, multi-agent interactive framework to generate long-context questions efficiently. Our approach efficiently generates high-quality single- and multi-page questions for extensive English and Arabic documents, covering hundreds of pages across diverse domains. This facilitates the development of LVLMs with enhanced long-context understanding ability. Experimental results in this work have shown that our generated English and Arabic questions (\textbf{AraEngLongBench}) are quite challenging to major open- and close-source LVLMs. The code and data proposed in this work can be found in \url{https://github.com/wangk0b/Multi_Agentic_QA_Long_Doc.git}. Sample Question and Answer (QA) pairs and structured system prompts can be found in the Appendix. 
\end{abstract}

\section{Introduction}
\label{sec:intro}
Document Understanding (DU) is a challenging task, especially for documents with complicated layouts and a lot of contextual dependencies. Large Vision-Language Models (LVLMs) have achieved plenty of progress on short-context DU tasks. Nevertheless, leading models such as OpenAI's GPT family \citep{achiam2023gpt}, Google's Gemini series \citep{gemini2024}, Anthropic's Claude series \citep{anthropic2024claude3}, and open-source models like InternLM-XC2-4KHD \citep{dong2024internlm}, LLaVA-NeXT \citep{li2024llava}, and CogVLM \citep{wang2023cogvlm} still face numerous challenges with multi-page documents. Despite their impressive performance on short-context benchmarks such as DocVQA \citep{mathew2021docvqa}, ChartQA \citep{masry2022chartqa}, InfographicVQA \citep{mathew2022infographicvqa}, and the tasks introduced in \cite{zhu2022towards}, accuracy rates remain at around 40\% even for long-context DU benchmarks such as DUDE \citep{van2023document}, MMLongBench \citep{ma2024mmlongbench}, LongDocURL \citep{deng2024longdocurl}, and M-longdoc \citep{chia2024m}. Even certain LVLMs even underperform purely text-based LLMs working on OCR-extracted content. This performance shortfall is primarily attributed to the weak capacity of LVLMs in picking up cross-page dependencies due to a scarcity of exposure to diverse, fine-grained, multi-page Q/A pairs throughout training. Absence of such information further aggravates for low-resource languages such as Arabic, significantly attributing the complexity of long-context DU tasks \citep{ghaboura2024camel}.

\begin{figure}
    \centering
    \begin{tikzpicture}
        \begin{axis}[
            width=0.85\linewidth,
            height=7cm,
            xlabel={Average Page Numbers},
            ylabel={Average Text Tokens},
            xmode=log,
            ymode=log,
            log basis x=10,
            log basis y=10,
            grid=both,
            enlargelimits=0.2,  % Adds extra space to prevent label cropping
            clip=false,  % Ensures labels are not clipped
            scatter/classes={
                arab={mark=*, fill =orange, draw=orange},
                normal={mark=*,fill = blue, draw=blue},
                ours={mark=star,draw=red}
            }
        ]
        
        % Plot normal points in blue
        \addplot[scatter, only marks, scatter src=explicit symbolic] 
        coordinates {
            (1, 100) [arab]
            (1, 151) [normal] 
            (1, 236) [normal] 
            (1, 7000) [normal] 
            (5.7, 1831) [normal] 
            (20, 2030) [normal] 
            (47.5, 21214) [normal] 
            (77, 84000) [ours]  % Ours is a red star
        };

        % Add labels with adjusted positions
        \node at (axis cs:1,100) [anchor=west, font=\small] {Camel};
        \node at (axis cs:1,151) [anchor=east, font=\small] {ChartQA};
        \node at (axis cs:1,236) [anchor=west, font=\small] {DocVQA};
        \node at (axis cs:1,7000) [anchor=north east, font=\small] {PWC};
        \node at (axis cs:5.7,1831) [anchor=north east, font=\small] {DUDE};
        \node at (axis cs:20,2030) [anchor=west, font=\small] {SlideVQA};
        \node at (axis cs:47.5,21214) [anchor=north west, font=\small, rotate=15] {MMLong};
        \node at (axis cs:50,84000) [anchor=south west, font=\small, red, rotate=15] {AraEngLongBench};

        \end{axis}
    \end{tikzpicture}
    \caption{Average Page Numbers vs. Average Text Tokens}
    \label{fig:avg_page_tokens}
\end{figure}
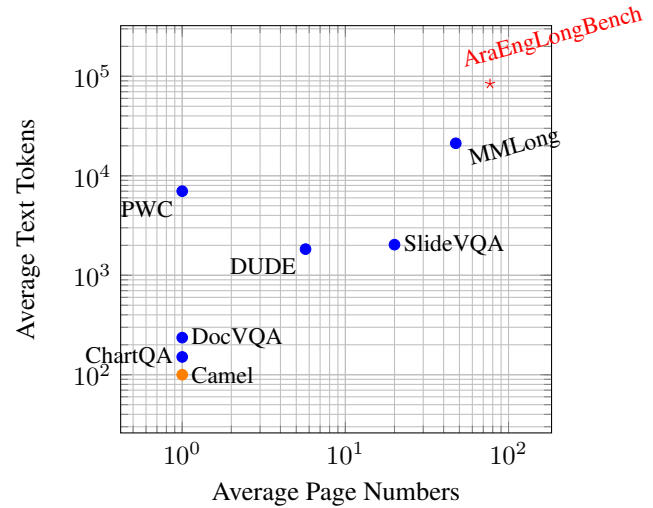
Although there is a boost in benchmarks for long-context document understanding (DU), i.e., DUDE, MMLongBench, LongDocURL, and M-longdoc, they rely primarily on human annotators and it is not easy to scale to large domains and languages \citep{chia2024m}. On low-resource languages like Arabic with restricted high-quality training data available, LVLM performance gets significantly impacted.

To meet these limitations, we introduce an independent, multi-agent framework for large-scale, long-context question generation. Based on a single prompt, our method applies virtual agents to collaborate in generating full single- and multi-page questions from large English and Arabic documents. Unlike manual approaches, this end-to-end process greatly enhances scalability and reduces annotation time.

\noindent In summary, our key contributions are as follows:
\begin{enumerate}[label=\arabic*. , leftmargin=0.3in]
     \item[1.] \textbf{We introduce a multi-agent interactive framework that generates high-quality single- and multi-page questions across domains (e.g., Chemistry, Biology, Math, Computer Science) from long documents, requiring only a single input prompt.}
    \item[2.] \textbf{We craft a standalone, agent-driven pipeline to gather sprawling Arabic documents in their entirety. In addition, our framework also handles question generation, directly tackling the data shortages that have long held back LVLMs in low-resource languages like Arabic.}
    \item[3.] \textbf{We run a battery of rigorous tests on our generated data; the results not only underscore the challenge these questions pose but also reveal clear gains in the models’ long-context understanding, proving that our automated pipeline can deliver data that meaningfully pushes the state of the art.}
\end{enumerate}
Overall, we believe this work significantly contributes to the DU community, particularly in long-context DU, and establishes a strong foundation for training more capable LVLMs for real-world applications. Figure~\ref{fig:avg_page_tokens} indicates that the documents we have in our dataset exceeds other datasets in both average length and average text tokens.

\section{Related Work}
\label{sec:formatting}
\subsection{DU Datasets} \label{R_BD}
Over the years, researchers have rolled out a host of datasets to tackle different facets of document understanding (DU)—from parsing page layouts \citep{pfitzmann2022doclaynet} to crafting concise summaries \citep{ghalandari2020large}, pinpointing key information \citep{vsimsa2023docile}, and even fielding visual question‐answering challenges \citep{mathew2020document}. Others dive into visual reasoning \citep{van2023document} or focus on making sense of tables and charts \citep{hu2024novachart}. 

However, most publicly available benchmarks remain tethered to narrow domains and single-page snapshots, which means they have not fully explored the adaptability of these vision-language giants to more complex layouts or multi-page narratives. It is true that a few newer benchmarks such as MMLongBench and M-LongDoc push into the multi-page arena \citep{ma2024mmlongbench,chia2024m}. Nonetheless, these benchmarks are still limited on document length, density of content, or the intensive effort of manual annotation. 

\subsection{LVLMs}

Document understanding (DU) systems generally fall into one of two camps. The first, often dubbed the \textbf{cascaded approach}, begins by running OCR to extract text, then treats the visual and linguistic streams separately with each modality encoded on its own before being fused downstream \citep{smith2007overview}. The second school of thought embraces an \textbf{end-to-end} vision-based paradigm, exemplified by the leading LVLMs. Here, models ingest the raw document image through a vision encoder such as CLIP \citep{radford2021clip}, BLIP-2 \citep{li2023blip2} or BRAVE \citep{kar2024brave} and directly align those visual features with word embeddings generated by a large language model.

\subsection{Agentic Workflow for Data Annotation}
Mass-scale LVLM or LLM training is infeasible with cost and time requirements due to the trillions of high-quality annotated data points required. Agentic workflow applications using autonomous or semi-autonomous AI agents provide scalable and cost-effective alternatives, continuously improving annotations while minimizing human intervention \citep{pangakis2023automated}. Nonetheless, existing state-of-the-art systems such as LabelVizier \citep{yang2023labelvizier} and MEGAnno+ \citep{kim2024meganno} remain highly relevant to human-agent collaboration. In contrast, our proposed multi-agent interactive setup offers end-to-end automatic solution, establishing a new standard for large-scale, high-quality annotation of long-context Document Understanding tasks.

\section{Agentic Workflow}

\subsection{Agentic Data Collection}
\begin{figure}[h!]
    \centering
    \includegraphics[width=1\linewidth]{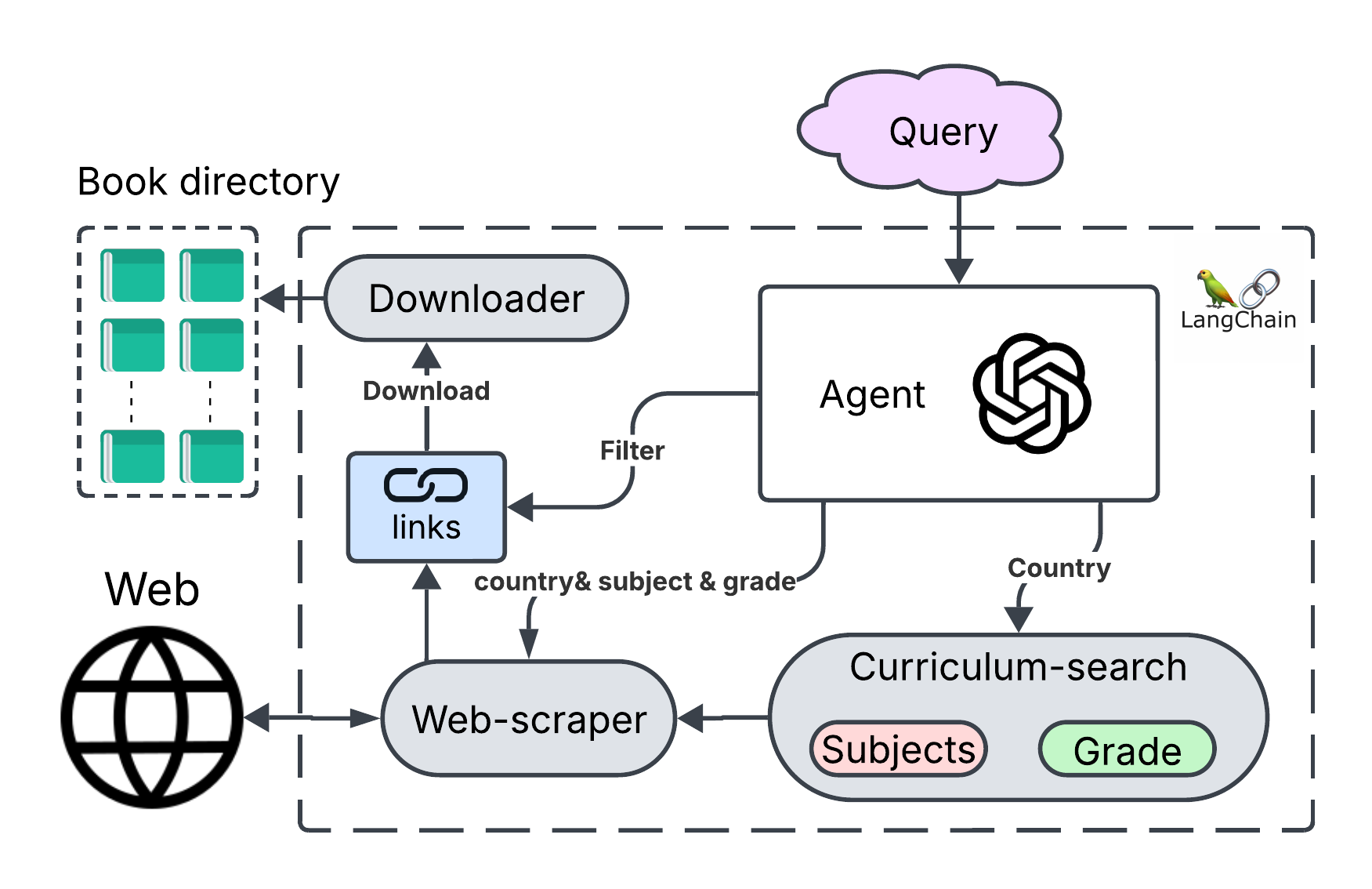}
    \caption{Agentic workflow for automated data collection.}
    \label{fig:workflow_scrapper}
\end{figure}

We used web scraping methods (see Figure~\ref{fig:workflow_scrapper}) to gather a large number of long-form documents from a variety of online sources. We systematically left out documents that failed to satisfy the minimum length requirements, possessed restrictive licenses, or defective in generating questions and answers. By doing so, we encountered a set of technical challenges, from multiple formats to failed extraction occurring from time to time and fetching incomplete content.

Adding Arabic texts was only more of the same. Handling right-to-left scripts and multiple character encodings often destroyed parsing pipelines. OCR accuracy plummeted for scanned or manually written documents. Arabic diacritic marks and dialect differences made normalization even harder. This made it harder to classify documents. In order to get structured content out of PDF and HTML documents, we have to implement parsers ourselves. This was especially challenging because we had to keep in mind that there were no conventions for standard metadata.

\subsection{Multi-Agent Interactive Question Generation Framework for Long Document Understanding}

This section formally presents the Multi-Agent Interactive Question Generation Framework for Long Document Understanding (Long DU) workflow. The objective of this workflow is to generate high-quality questions and answers that are contextualized from long documents with complex structure. The framework utilizes a multi-agent system which employs an iterative process for the generation, validation, and refinement of questions which are contextual.
\begin{figure*}[t]
    \centering
    \includegraphics[width=1\linewidth]{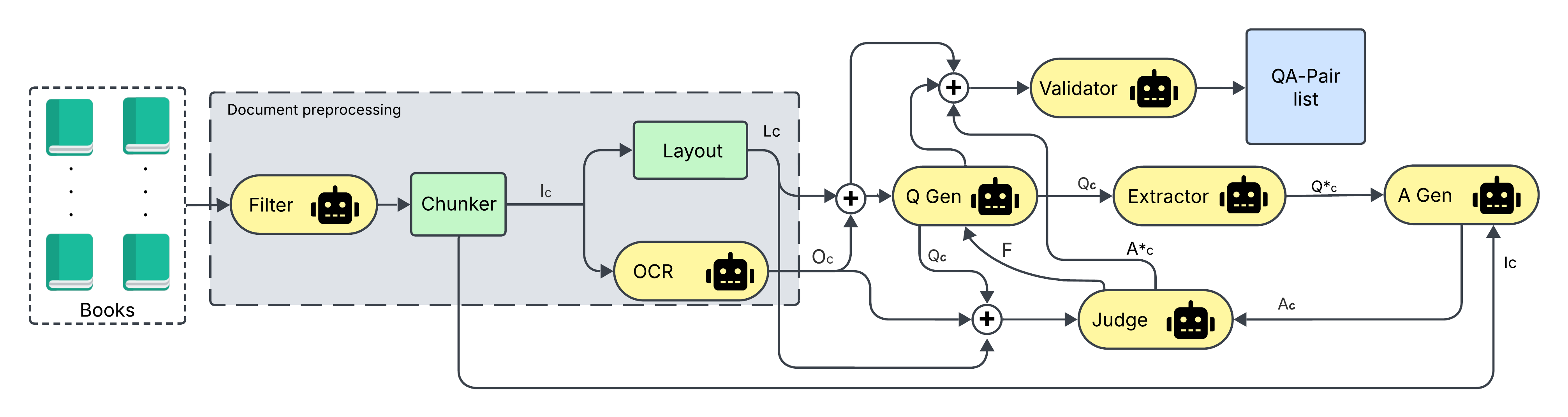}
    \caption{Agentic workflow for automated QA generation.}
    \label{fig:workflow}
\end{figure*}

\subsubsection{Document Preprocessing}

The workflow begins with document preprocessing, transforming input PDF documents into structured representations that facilitate further processing. The key steps include:

The proposed framework first reads input documents in PDF format. Each PDF page is then converted to an image representation ($I$) with the aid of the \texttt{pdf2image} library \citep{belval2018pdf2image}, thereby preserving the original document structure.

Second, OCR is applied to the images to extract text information ($O$). Instead of employing a generic fixed OCR engine, the system employs a large vision-language model (LVLM)-based OCR method \citep{zhang2024texthawk2,liu2024oceanocr}. At the same time, the system employs a deep learning-based layout analysis model, for instance, YOLO \citep{YOLOReview2024}, to identify structural elements, including headings, paragraphs, tables, and figures, to facilitate proper segmentation of document components ($L$).

Then the documents are split into processible pieces ($I_c$), with a 10-page overlap for coherence of context in downstream processing by LVLMs. For every processible piece, corresponding OCR and layout analysis outputs ($O_c$, $L_c$) are fetched, providing structured input to facilitate effective question generation.

\subsubsection{Multi-Agent Interactive Chain}

Once the preprocessing is done, the system applies a chain of expert agents in an iterative and synergistic manner to generate and refine questions. First, \textbf{Agent 1 (Question Generation)} produces high-quality questions ($Q_c$) based on policy $\pi$, utilizing OCR outputs ($O_c$) and layout analysis ($L_c$) to maintain document content and evidence in sync.

Then, \textbf{Agent 2 (Question Extraction)} utilizes these generated questions ($Q_c$) and filtered questions ($Q^*_c$) to remove irrelevant or redundant information. Finally, \textbf{Agent 3 (Answer Generation)} utilizes the chunked image segments ($I_c$) with filtered questions ($Q^*_c$) to produce candidate answers ($A_c$) that are directly grounded in the content of the document.

To analyze and enhance the quality of these outputs,
\textbf{Agent 4 (Assessment and Feedback)} aggregates all available context, including image segments ($I_c$), OCR outputs ($O_c$), layout information ($L_c$), synthesized questions ($Q_c$), candidate answers ($A_c$), and pre-defined policies ($\pi$). The agent produces reference answers ($A^*_c$), estimates depth and difficulty for each question, and provides iterative feedback ($F$) to guide subsequent improvements.

Trained with this feedback ($F$), \textbf{Agent 1 (Question Refinement)} refines and revises the initial question set ($Q_c$), iteratively performing Steps 1 to 5 until attaining the target level of complexity and quality. Additionally, \textbf{Agent 5 (Evidence Validation)} performs an end-of-process evaluation based on chunked images ($I_c$), OCR results ($O_c$), layout analysis ($L_c$), recommended questions ($Q_c$), and reference answers ($A^*_c$), confirming consistency and coherence among generated answers and supporting evidence. These rigorous validation questions ($\hat{Q}_c$) thus created are then finalized. 

Notably, if \textbf{Agent 4} detects an accuracy rate exceeding 40\%, it triggers \textbf{Agent 1} to increase question complexity, challenging the answer generation agent. \textbf{Agent 5} ensures final question-answer pairs align with validated sources, mitigating mismatches observed in prior benchmarks \citep{ma2024mmlongbench}.

\section{Data Analysis}
The final dataset consists of well-structured tuples containing (question, answer, evidence pages, evidence sources, justification, and validation), making it a robust resource for long-document understanding. Here is the detailed statistics: \textbf{4160 (Reasoning)} + \textbf{271 (Factual Recall)} + \textbf{412 (Image-based Question)} + \textbf{15 (Prediction Analysis)} + \textbf{36 (Step-by-step Explanation)} + \textbf{272 (Conceptual Understanding)} + \textbf{798 (Hypothetical Reasoning)} + \textbf{281 (Multi-hop Reasoning)} + \textbf{6 (Data Retrieval \& OCR)} + \textbf{21 (Experimental Design)} + \textbf{6 (Argumentation)} + \textbf{454 (Unanswerable)} = \textbf{6732} questions. 

The most frequent question type is \textit{Reasoning}, with 4,160 instances (61.8\%), emphasizing deductive ability, inference, and problem-solving. The second most frequent question type is \textit{Hypothetical Reasoning} (798 instances, 11.9\%), with high-level reasoning scenarios. Minor categories include \textit{Unanswerable} (6.7\%), \textit{Image-based} (6.1\%), \textit{Multi-hop reasoning} (4.2\%), \textit{Conceptual Understanding} (4.0\%), and \textit{Factual Recall} (4.0\%), that together measure different cognitive abilities such as anticipating consequences, comprehension of visuals, and conceptual knowledge. The most underrepresented categories—\textit{Argumentation}, \textit{Data Retrieval \& OCR}, and \textit{Experimental Design} (all below 0.5\%)—indicate minimal emphasis on argument construction, data retrieval, and experiment design.

\begin{table*}[h!]
    \centering
    \renewcommand{\arraystretch}{1.2} % Adjust row height
    \setlength{\tabcolsep}{4pt} % Adjust column spacing
     \resizebox{\textwidth}{!}{  
    \begin{tabular}{lcc|cccccc|cccccc|l}
        \toprule
        \textbf{Model} & \textbf{Param} & \textbf{CW} 
        & \multicolumn{6}{c|}{\textbf{English}} 
        & \multicolumn{6}{c|}{\textbf{Arabic}} 
        & \textbf{Average} \\
        \cmidrule(lr){4-9} \cmidrule(lr){10-15}
        & & & SC & MC & LC & SP & CP & UA 
        & SC & MC & LC & SP & CP & UA 
        &  \\
        \midrule
        \multicolumn{15}{l}{\textbf{Closed-Source Models}} \\
        \midrule
        GPT-4o & -- & 128K  & 44.8\%& 36.1\%& 27.8\%& 50.5\%& 31.4\%&18.1\%&32.9\%& 31.0\%&23.1\%& 46.0\%& 21.5\%& 12.5\% & 29.8\%  \\
        Gemini-2.0 Flash & -- & 1M  &49.7\% & 56.5\% & 56.2\%&72.2\%&71.8\%&10.2\%&44.2\%& 45.1\%& 35.4\%&47.2\%&53.2\%& 8.6\% & 49.6\%  \\
        Gemini-1.5 Pro  & -- & 2M &49.6\%&55.1\%&58.2\%&73.9\%&73.7\%& 8.9\%&42.6\%&42.8\%&42.3\%&51.6\%& 49.8\%&8.8\% & 49.8\% \\
        \midrule
        \multicolumn{15}{l}{\textbf{Open-Source Models}} \\
        \midrule
        Qwen 2 VL & 7B & 32K &49.4\%&53.7\%&52.3\%&67.7\%&69.4\%&7.1\%&29.6\%&21.2\%&9.9\%&27.6\%&35.9\%&5.1\%&38.4\%\\
        Qwen 2.5 VL & 7B & 128K &49.9\%&54.6\%&54.9\%&70.6\%&71.7\%&7.1\%&37.6\%&32.4\%&27.2\%&41.8\%&41.7\%&5.9\%& 44.5\% \\
        Llama 3.2 VL & 11B & 128K & 32.4 \% & NA & NA&37.3\%& 53.3\%& 3.2\% &NA & NA &NA&NA&NA&NA& 32.4\% \\
        \bottomrule
    \end{tabular}
    }
    \caption{Comparison of models based on parameters, context window, and language capabilities. CW = context window, SC = short-context, MC = medium-context, LC = long-context, SP = Single-Page, CP = Cross-Page, UA = Unanswerable.}
    \label{tab:model_comparison}
\end{table*}

After assigning \textbf{5085} answers to the filtered questions into eight categories, the answer types are distributed as follows; the most frequent type is \textbf{Text} with \textbf{2826} answers (approximately 55.6\%). The second most frequent type is \textbf{Integer} with \textbf{1169} occurrences (approximately 23.0\%), indicating a high percentage of questions anticipating whole numbers as answers. \textbf{Code} ranks in third place with \textbf{624} answers (approximately 12.3\%), which can be seen to represent programming effort. \textbf{Float} response occupies \textbf{347} (approximately 6.8\%). Other, less common structures are \textbf{List} (\textbf{74}), \textbf{Boolean} (\textbf{31}), \textbf{Array} (\textbf{15}), and \textbf{JSON} (\textbf{7}). The division signals a collection generated to exercise with a broad span of output type, from unstructured natural language text through structured numerical, computer, and data forms.

\section{Experiments}

Table~\ref{tab:model_comparison} presents a comparative performance of open-source and closed-source Vision-Language Large Models (V-LLMs) in English and Arabic benchmarks. These measurements capture the model's performance in different contextual environments such as short-context (SC: $<100$), medium-context (MC: $100-200$), long-context (LC: $>200$), single-page (SP), cross-page (CP), and unanswerable (UA) types of queries. Performance of the model is presented in percentage accuracy for conveying insights into their performance based on task difficulty and language.

Among closed-source models, Gemini-1.5 Pro has the highest overall average score (49.8\%), followed closely by Gemini-2.0 Flash (49.6\%). Both Gemini models perform very well on nearly all English context lengths and types of reasoning, especially in SP and CP categories—indicating high-level contextual and visual reasoning abilities. However, their accuracy on unanswerable (UA) questions remains fairly low (~9\%), indicating a continued problem in having the ability to abstain confidently when one should.

GPT-4o, marginally lower in mean score compared to the Gemini models (29.8\%), is evenly well-performing in English and Arabic benchmarks with exceptional performance in SP for English (50.5\%). Its handling of UA is comparatively stronger (18.1\% English, 12.5\% Arabic), suggesting a more cautious approach towards uncertain conditions.

Among open-source systems, Qwen 2.5 VL emerges as the leader with a mean of 44.5\%, outperforming its earlier system Qwen 2 VL by over 6\%. It achieves high consistency for English SC–CP tasks and shows improved performance in Arabic, especially under LC and CP environments. Its accuracy, however, drops very markedly on UA questions (5.9\%), demonstrating low resilience to vague or adversarial questions.

LLaMA 3.2 VL does considerably worse with partial evaluation (NA in the majority of categories) and it is difficult to state conclusively. However, its low reported statistics (e.g., 32.4\% in SC English) suggest unripe abilities, particularly in multilingual or rich settings.

\section{Conclusion}
In this paper, we have developed a Multi-Agent Interactive Question Generation Framework specifically designed for long-context Document Understanding (DU). The automated system effectively generates high-quality, contextually-relevant question-answer pairs from long documents. Leverage on advanced preprocessing techniques like document layout analysis, OCR, and content segmentation, the system stringently preprocesses documents to facilitate accurate information extraction.

Our multi-agent system involves agents that operate in a cooperative iterative feedback loop: one generates initial questions, another refines them, a third provides answers, a fourth assesses question quality, and a fifth tests for consistency and correctness of supporting evidence. Such a structured, programmatic procedure significantly advances the state of Large Vision-Language Model (LVLM) techniques by providing diverse and informative training data while enabling clear examination of model behavior.

Experiments on our proposed \textbf{AraEngLongBench} dataset reveal that closed-source models such as Gemini-1.5 Pro are currently state-of-the-art. However, open-source models such as Qwen 2.5 VL are catching up rapidly, particularly for English tasks. What is notable is that all the tested models underperform for Arabic understanding tasks and struggle with correctly labeling unanswerable questions with confident tags. Cross-lingual generalization and confidence calibration are important avenues of future work to improve LVLM robustness and fairness.

In addition, in future work, we will explore how depth of document understanding is related to the number of iterative agent loops. We also plan to further fine-tune the proposed framework for varying applications and further enhance overall processing efficiency.


\bibliographystyle{IEEEbib}
\bibliography{refs}

\begin{thebibliography}{10}

\bibitem{achiam2023gpt}
Josh Achiam, Steven Adler, Sandhini Agarwal, Lama Ahmad, Ilge Akkaya, Florencia~Leoni Aleman, Diogo Almeida, Janko Altenschmidt, Sam Altman, Shyamal Anadkat, et~al.,
\newblock ``Gpt-4 technical report,''
\newblock {\em arXiv preprint arXiv:2303.08774}, 2023.

\bibitem{gemini2024}
{Gemini Team},
\newblock ``Gemini 1.5: Unlocking multimodal understanding across millions of tokens of context,'' 2024,
\newblock Accessed: 2025-02-11.

\bibitem{anthropic2024claude3}
Anthropic,
\newblock ``Claude 3 haiku: Our fastest model yet,'' 2024,
\newblock Accessed: 2025-02-11.

\bibitem{dong2024internlm}
Xiaoyi Dong, Pan Zhang, Yuhang Zang, Yuhang Cao, Bin Wang, Linke Ouyang, Songyang Zhang, Haodong Duan, Wenwei Zhang, Yining Li, et~al.,
\newblock ``Internlm-xcomposer2-4khd: A pioneering large vision-language model handling resolutions from 336 pixels to 4k hd,''
\newblock {\em arXiv preprint arXiv:2404.06512}, 2024.

\bibitem{li2024llava}
Bo~Li, Kaichen Zhang, Hao Zhang, Dong Guo, Renrui Zhang, Feng Li, Yuanhan Zhang, Ziwei Liu, and Chunyuan Li,
\newblock ``Llava-next: Stronger llms supercharge multimodal capabilities in the wild,'' 2024.

\bibitem{wang2023cogvlm}
Weihan Wang, Qingsong Lv, Wenmeng Yu, Wenyi Hong, Ji~Qi, Yan Wang, Junhui Ji, Zhuoyi Yang, Lei Zhao, Xixuan Song, et~al.,
\newblock ``Cogvlm: Visual expert for pretrained language models,''
\newblock {\em arXiv preprint arXiv:2311.03079}, 2023.

\bibitem{mathew2021docvqa}
Minesh Mathew, Dimosthenis Karatzas, and CV~Jawahar,
\newblock ``Docvqa: A dataset for vqa on document images,''
\newblock in {\em Proceedings of the IEEE/CVF winter conference on applications of computer vision}, 2021, pp. 2200--2209.

\bibitem{masry2022chartqa}
Ahmed Masry, Do~Xuan Long, Jia~Qing Tan, Shafiq Joty, and Enamul Hoque,
\newblock ``Chartqa: A benchmark for question answering about charts with visual and logical reasoning,''
\newblock {\em arXiv preprint arXiv:2203.10244}, 2022.

\bibitem{mathew2022infographicvqa}
Minesh Mathew, Viraj Bagal, Rub{\`e}n Tito, Dimosthenis Karatzas, Ernest Valveny, and CV~Jawahar,
\newblock ``Infographicvqa,''
\newblock in {\em Proceedings of the IEEE/CVF Winter Conference on Applications of Computer Vision}, 2022, pp. 1697--1706.

\bibitem{zhu2022towards}
Fengbin Zhu, Wenqiang Lei, Fuli Feng, Chao Wang, Haozhou Zhang, and Tat-Seng Chua,
\newblock ``Towards complex document understanding by discrete reasoning,''
\newblock in {\em Proceedings of the 30th ACM International Conference on Multimedia}, 2022, pp. 4857--4866.

\bibitem{van2023document}
Jordy Van~Landeghem, Rub{\`e}n Tito, {\L}ukasz Borchmann, Micha{\l} Pietruszka, Pawel Joziak, Rafal Powalski, Dawid Jurkiewicz, Micka{\"e}l Coustaty, Bertrand Anckaert, Ernest Valveny, et~al.,
\newblock ``Document understanding dataset and evaluation (dude),''
\newblock in {\em Proceedings of the IEEE/CVF International Conference on Computer Vision}, 2023, pp. 19528--19540.

\bibitem{ma2024mmlongbench}
Yubo Ma, Yuhang Zang, Liangyu Chen, Meiqi Chen, Yizhu Jiao, Xinze Li, Xinyuan Lu, Ziyu Liu, Yan Ma, Xiaoyi Dong, et~al.,
\newblock ``Mmlongbench-doc: Benchmarking long-context document understanding with visualizations,''
\newblock {\em arXiv preprint arXiv:2407.01523}, 2024.

\bibitem{deng2024longdocurl}
Chao Deng, Jiale Yuan, Pi~Bu, Peijie Wang, Zhong-Zhi Li, Jian Xu, Xiao-Hui Li, Yuan Gao, Jun Song, Bo~Zheng, et~al.,
\newblock ``Longdocurl: a comprehensive multimodal long document benchmark integrating understanding, reasoning, and locating,''
\newblock {\em arXiv preprint arXiv:2412.18424}, 2024.

\bibitem{chia2024m}
Yew~Ken Chia, Liying Cheng, Hou~Pong Chan, Chaoqun Liu, Maojia Song, Sharifah~Mahani Aljunied, Soujanya Poria, and Lidong Bing,
\newblock ``M-longdoc: A benchmark for multimodal super-long document understanding and a retrieval-aware tuning framework,''
\newblock {\em arXiv preprint arXiv:2411.06176}, 2024.

\bibitem{ghaboura2024camel}
Sara Ghaboura, Ahmed Heakl, Omkar Thawakar, Ali Alharthi, Ines Riahi, Abduljalil Saif, Jorma Laaksonen, Fahad~S Khan, Salman Khan, and Rao~M Anwer,
\newblock ``Camel-bench: A comprehensive arabic lmm benchmark,''
\newblock {\em arXiv preprint arXiv:2410.18976}, 2024.

\bibitem{pfitzmann2022doclaynet}
Birgit Pfitzmann, Christoph Auer, Michele Dolfi, Ahmed~S Nassar, and Peter Staar,
\newblock ``Doclaynet: A large human-annotated dataset for document-layout segmentation,''
\newblock in {\em Proceedings of the 28th ACM SIGKDD conference on knowledge discovery and data mining}, 2022, pp. 3743--3751.

\bibitem{ghalandari2020large}
Demian~Gholipour Ghalandari, Chris Hokamp, Nghia~The Pham, John Glover, and Georgiana Ifrim,
\newblock ``A large-scale multi-document summarization dataset from the wikipedia current events portal,''
\newblock {\em arXiv preprint arXiv:2005.10070}, 2020.

\bibitem{vsimsa2023docile}
{\v{S}}t{\v{e}}p{\'a}n {\v{S}}imsa, Milan {\v{S}}ulc, Michal U{\v{r}}i{\v{c}}{\'a}{\v{r}}, Yash Patel, Ahmed Hamdi, Mat{\v{e}}j Koci{\'a}n, Maty{\'a}{\v{s}} Skalick{\`y}, Ji{\v{r}}{\'\i} Matas, Antoine Doucet, Micka{\"e}l Coustaty, et~al.,
\newblock ``Docile benchmark for document information localization and extraction,''
\newblock in {\em International Conference on Document Analysis and Recognition}. Springer, 2023, pp. 147--166.

\bibitem{mathew2020document}
Minesh Mathew, Ruben Tito, Dimosthenis Karatzas, R~Manmatha, and CV~Jawahar,
\newblock ``Document visual question answering challenge 2020,''
\newblock {\em arXiv preprint arXiv:2008.08899}, 2020.

\bibitem{hu2024novachart}
Linmei Hu, Duokang Wang, Yiming Pan, Jifan Yu, Yingxia Shao, Chong Feng, and Liqiang Nie,
\newblock ``Novachart: A large-scale dataset towards chart understanding and generation of multimodal large language models,''
\newblock in {\em Proceedings of the 32nd ACM International Conference on Multimedia}, 2024, pp. 3917--3925.

\bibitem{smith2007overview}
Ray Smith,
\newblock ``An overview of the tesseract ocr engine,''
\newblock in {\em Ninth international conference on document analysis and recognition (ICDAR 2007)}. IEEE, 2007, vol.~2, pp. 629--633.

\bibitem{radford2021clip}
Alec Radford, Jong~Wook Kim, Chris Hallacy, Aditya Ramesh, Gabriel Goh, Sandhini Agarwal, et~al.,
\newblock ``Learning transferable visual models from natural language supervision,''
\newblock {\em arXiv preprint arXiv:2103.00020}, 2021.

\bibitem{li2023blip2}
Junnan Li, Dongxu Li, Silvio Savarese, and Steven Hoi,
\newblock ``Blip-2: Bootstrapping language-image pre-training with frozen image encoders and large language models,''
\newblock {\em arXiv preprint arXiv:2301.12597}, 2023.

\bibitem{kar2024brave}
Omid~Faraji Kar, Alessio Tonioni, Petra Poklukar, Anshul Kulshrestha, Amir Zamir, and Federico Tombari,
\newblock ``Brave: Broadening the visual encoding of vision-language models,''
\newblock {\em arXiv preprint arXiv:2404.07204}, 2024.

\bibitem{pangakis2023automated}
N.~Pangakis, S.~Wolken, and N.~Fasching,
\newblock ``Automated annotation with generative ai requires validation,''
\newblock {\em arXiv preprint}, 2023.

\bibitem{yang2023labelvizier}
Xinyi Yang, Lei Zhang, Saurabh Trivedi, Qian Liao, Byron~C. Wallace, and Matthew Lease,
\newblock ``Labelvizier: Error profiling and interactive data annotation for long-document understanding,''
\newblock {\em arXiv preprint}, 2023.

\bibitem{kim2024meganno}
H.~Kim, K.~Mitra, R.~L. Chen, S.~Rahman, and D.~Zhang,
\newblock ``Meganno+: A human-llm collaborative annotation system,''
\newblock {\em arXiv preprint}, 2024.

\bibitem{belval2018pdf2image}
Jeremy Belval,
\newblock ``pdf2image: A python library to convert pdf pages to images using poppler,'' GitHub repository, 2018.

\bibitem{zhang2024texthawk2}
Xinyu Zhang, Lin Zhang, Jiaxin Wang, et~al.,
\newblock ``Texthawk2: A large vision-language model excels in bilingual ocr and grounding with 16x fewer tokens,''
\newblock {\em arXiv preprint arXiv:2410.05261}, 2024.

\bibitem{liu2024oceanocr}
Zhaoyi Liu, Yifan Li, Qi~Wang, et~al.,
\newblock ``Towards general ocr application via a vision-language model,''
\newblock {\em arXiv preprint arXiv:2501.15558}, 2024.

\bibitem{YOLOReview2024}
Author(s),
\newblock ``The yolo framework: A comprehensive review of evolution and applications,''
\newblock {\em Computers}, vol. 13, no. 12, pp. 336, 2024.

\end{thebibliography}

\end{document}